%% file: main.tex
\PassOptionsToPackage{prologue,dvipsnames}{xcolor}
\documentclass[sigconf, nonacm]{acmart}

\newcommand\vldbdoi{XX.XX/XXX.XX}

\usepackage{amssymb}
\usepackage{amsmath} 
\usepackage{booktabs} 
\usepackage{array}
\usepackage{graphicx} 
\usepackage{multirow}
\usepackage{color}
\usepackage{makecell} 
\usepackage{float}
\usepackage{subcaption}
\usepackage{colortbl}
\usepackage{placeins}
\usepackage{comment}
\usepackage{pifont}
\newcommand{\cmark}{\ding{51}} 
\newcommand{\xmark}{\ding{55}} 

\usepackage[linesnumbered,vlined,ruled,noend]{algorithm2e}
\usepackage{algorithm2e}
\usepackage{amssymb}

\begin{document}
\title{Synth-Empathy: Towards High-Quality Synthetic Empathy Data}

\author{Hao Liang$^{\dagger\ddagger}$, Linzhuang Sun$^{\dagger\diamondsuit}$, Jingxuan Wei$^{\diamondsuit}$, Xijie Huang$^\ddagger$, Linkun Sun$^{\clubsuit}$, Bihui Yu$^{\diamondsuit}$, Conghui He$^{\spadesuit}$, Wentao Zhang$^{\ddagger*}$}
\affiliation{
$^\ddagger$Peking University~~~~~$^\diamondsuit$University of Chinese Academy of Sciences~~~~~$^{\spadesuit}$Shanghai AI Laboratory \newline $^\clubsuit$ Institute of Information Engineering, Chinese Academy of Sciences
}
\affiliation{
 $^\dagger$hao.liang@stu.pku.edu.cn, $^\dagger$sunlinzhuang21@mails.ucas.ac.cn,  wentao.zhang@pku.edu.cn
}

\begin{abstract}
In recent years, with the rapid advancements in large language models (LLMs), achieving excellent empathetic response capabilities has become a crucial prerequisite. Consequently, managing and understanding empathetic datasets have gained increasing significance. However, empathetic data are typically human-labeled, leading to insufficient datasets and wasted human labor. In this work, we present Synth-Empathy, an LLM-based data generation and quality and diversity selection pipeline that automatically generates high-quality empathetic data while discarding low-quality data. With the data generated from a low empathetic model, we are able to further improve empathetic response performance and achieve state-of-the-art (SoTA) results across multiple benchmarks. Moreover, our model achieves SoTA performance on various human evaluation benchmarks, demonstrating its effectiveness and robustness in real-world applications. Furthermore, we show the trade-off between data quantity and quality, providing insights into empathetic data generation and selection. The codebase and data are made available at \url{https://github.com/Aurora-slz/Synth-Empathy}.
\end{abstract}

\maketitle
\begingroup
\renewcommand\thefootnote{}\footnote{\noindent
$\dagger$ The first two authors have equal contributions. \\
$*$ Corresponding Author
}
\addtocounter{footnote}{-1}
\endgroup

%

\section{Introduction}
In recent years, with the rapid advancements in large language models (LLMs)~\cite{chatgpt, llama}, data management has become a crucial aspect of these technologies~\cite{fernandez2023large, trummer2023bert, chen2023lingua, miao2024demystifying, nie2023flexmoe}. At the same time, \citet{bai2024survey} also demonstrates that data processing, selection, and management can significantly influence the performance of LLMs.

Empathy, the ability to understand and share the feelings of another, is a critical component of human social interaction and communication. It allows individuals to connect with others on an emotional level, fostering relationships and promoting prosocial behavior ~\cite{davis1983measuring}. In the future of Human-central Artificial General Intelligence (AGI), excellent empathetic response capability is a crucial prerequisite, which focuses on equipping LLMs to understand and respond appropriately to human feelings ~\cite{rashkin2018towards}.

\begin{figure}
\centering 
\includegraphics[width=0.5\textwidth]{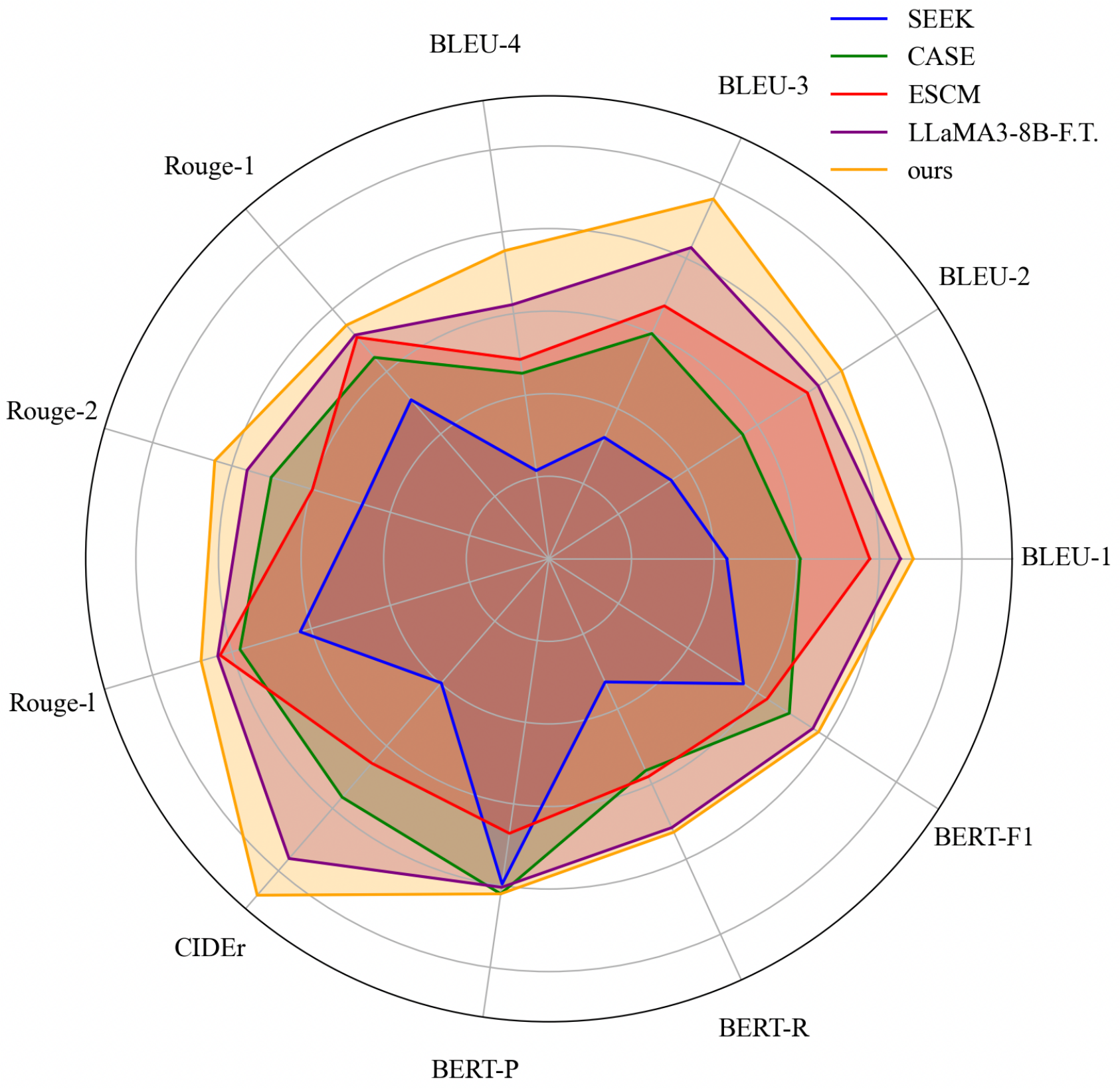} 
\caption{Comparison of our Synth-Empathy data-trained model with previous SoTA models. The results demonstrate that our model achieves superior performance on multiple empathetic benchmarks.}
\label{Fig.method}
\end{figure}
In previous research, empathy modeling has predominantly concentrated on human-labeled data ~\cite{wang2022empathetic, fu2023core, yang-etal-2023-exploiting-emotion, yufeng2024ctsm} and integrating information through modifications to the model architecture ~\cite{ghosal2020cosmic, zhou2021probing, sabour2022cem}. While the importance of model architectures is well acknowledged, the quality and quantity of data are also critical ~\cite{chen2023alpagasus, xu2023rethinking}. Additionally, \citet{fernandez2023large} indicates that new data management methods are needed for LLMs due to their massive data requirements. Previous empathetic studies underscore the significance of data-centric approaches. They frequently overlook effective data generation approaches, leading to the following two key challenges:


\textbf{C1. High Cost of Human-Labor.} 
Previous methods rely on human labor for the creation of empathetic datasets~\cite{wang2022empathetic, fu2023core, yang-etal-2023-exploiting-emotion, yufeng2024ctsm}. However, this process can be extremely costly and requires substantial human effort.

\textbf{C2. Poor Effectiveness.}
Previous research~\cite{lin2019moel, ghosal2020cosmic, inproceedings, sabour2022cem, wang2022empathetic, zhou2022case} has only a limited amount of data. Despite efforts by ~\citet{sun2024efficient} to improve model performance, the empathetic effectiveness remains constrained. More empathetic data is needed to train high-performance empathetic models.

\begin{figure}
\centering 
\includegraphics[width=0.5\textwidth]{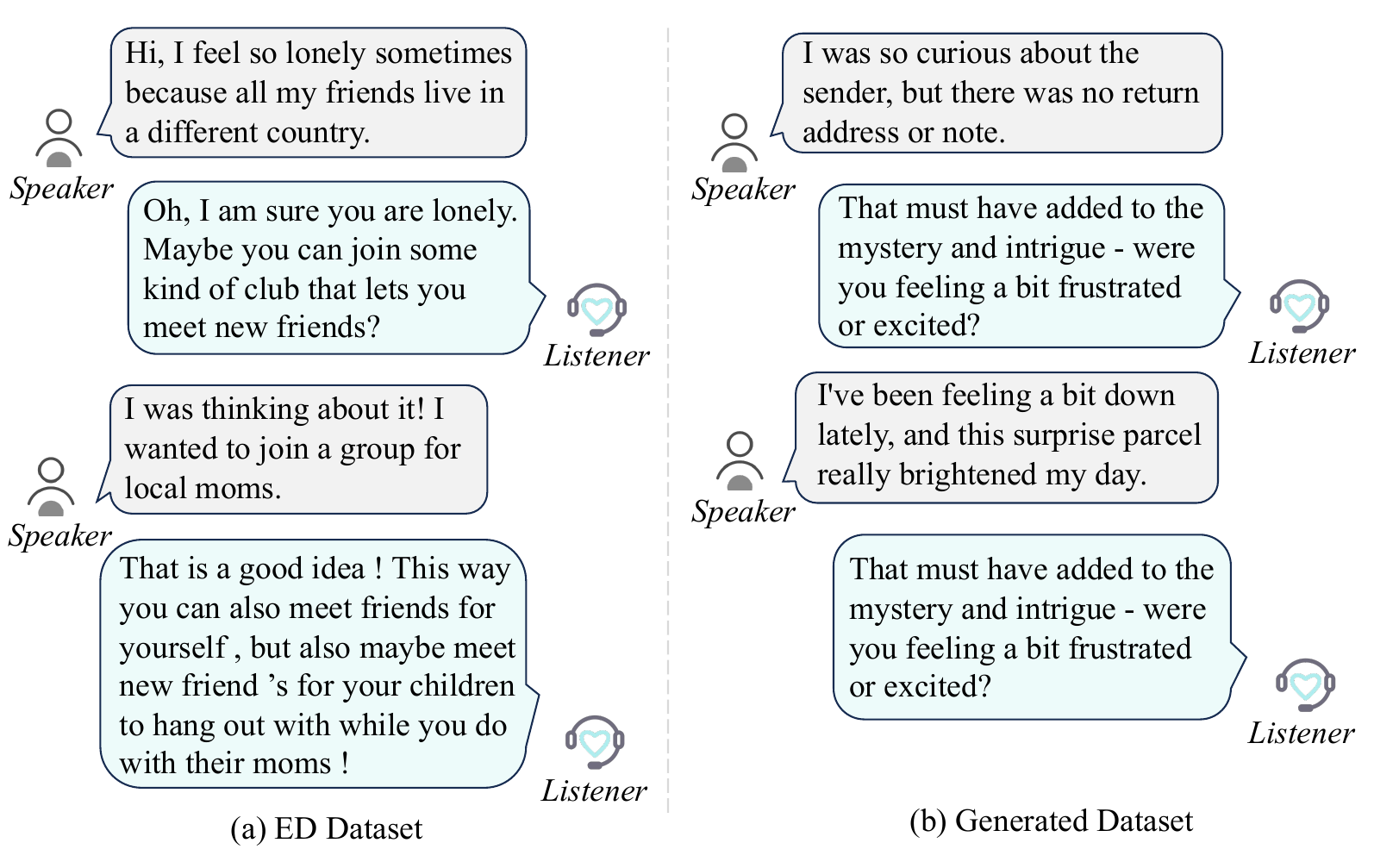} 
\caption{Comparison of Data Examples. (a) An example from the ED dataset. (b) An example from the synthetic dataset.}
\label{Fig.demo}
\end{figure}

To address these issues, as shown in Figure \ref{Fig.method}, we propose a three-step empathetic data generation and curation pipeline and subsequently train an empathetic LLM. First, we utilize prompts to generate empathetic responses. Next, we apply domain knowledge for quality selection to curate the empathetic data. Finally, we conduct diversity selection for further data curation. With the curated synthetic dataset, we fine-tune an LLM and achieve state-of-the-art (SoTA) empathetic response performance.

The core contributions of this paper are summarized as follows:

\begin{itemize}
\item \textbf{New Perspective.}
Limited data and low effectiveness are significant impediments to the practical adoption of empathy models. To the best of our knowledge, this study represents the first approach to generate data from scratch to address these challenges.
\item \textbf{New Method.}
We propose a new data generation and curation pipeline for empathy, introducing the first generated high-quality empathy dataset. Utilizing our meticulously curated synthetic dataset, we pioneer the integration of synthetic data to empathetic model training. Our method, which utilized curated synthetic data, enables robust effective, and user-friendly empathetic response.
\item \textbf{SoTA Performance.} 
    \textbf{(1) \textit{Effectiveness in Empathetic Response.}} 
    By utilizing the synthetic empathetic response data to fine-tune LLMs, our method can achieve SoTA performance in multiple benchmarks, as shown in Figure \ref{Fig.method}. At the same time, we achieve SoTA performance on human evaluation benchmarks. This means our model have high application potential. \\
    \textbf{(2) \textit{No Human Labor.}} 
    By utilizing curated prompts and LLMs, we can obtain high-quality empathetic responses without the need for human labor. With a meticulously designed data curation pipeline, as illustrated in Figure \ref{Fig.method}, we achieve high-quality data, as demonstrated in Figure \ref{Fig.demo} and Figure \ref{Fig.data-statistical}. This approach significantly reduces the requirement for human labor.
\end{itemize}
\section{Related Work}
\begin{figure*}[ht]
\centering 
\includegraphics[width=1.0\textwidth]{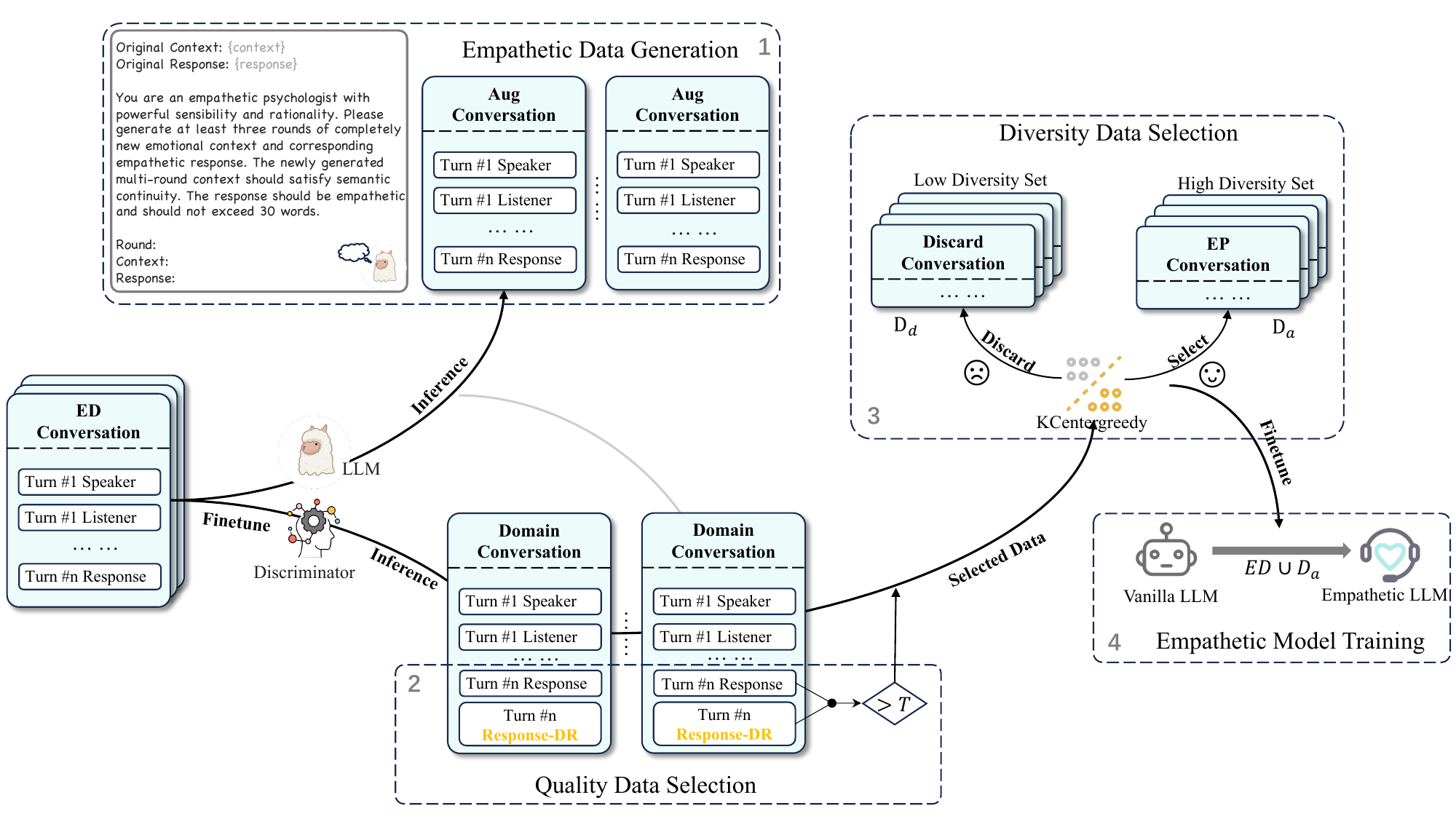} 
\caption{Empathetic Data Generation and Curation Pipeline, which is composed by (1) Empathetic Data Generation module, (2) Quality Data Selection module, (3) Diversity Data Selection module and (4) Empathetic Model Training module.}
\label{Fig.method}
\end{figure*}
\subsection{Empathetic Response Generation}
 Imbuing models with empathy to enhance emotional experience is a crucial aspect of AGI. ~\citet{rashkin2018towards} constructed the ED dataset, the most authoritative dataset for measuring the empathic abilities of models, which has since become a benchmark for empathy research. Building on this dataset, numerous researchers have focused on improving models' empathy comprehension by integrating emotional labels and external knowledge.

~\citet{sabour2022cem} introduces commonsense knowledge inference into this task by enriching historical conversation data with the pre-trained COEMT model~\cite{bosselut2019comet}. Diverged from this approach, ~\citet{li2022knowledge} incorporate external knowledge through Graph Neural Networks (GNN) for context encoding. Additionally, \citet{wang2022empathetic} posit that using detailed sentiment labels can improve the accuracy of capturing user sentiment and puts forward a sentiment loss mechanism with multiple levels of granularity to enhance model training.  Moreover, ~\citet{kim2022emp} strives to provide empathetic responses by analyzing context at the word level. ~\citet{chen2022wish} employed a emotion detection algorithm based on psychological principles to pinpoint important statements in conversations. ~\citet{zhao2022don} take a different approach by not only recognizing the emotions of others but also assessing saved model emotional state.  Furthermore, ~\citet{qian2023think} divides the empathetic response task into two steps: verifying the semantic content and infusing emotional expression.  

In the LLM era, many researchs explore ways to boost the empathic potential of models through meticulous design of prompts~\cite{qian2023harnessing, wang2023enhancing, yang2024iterative}. Differently, ~\citet{sun2023rational} delves into the general sensibility and rationality, exploring their respective contributions to empathy. However, the importance of fine-grained sensibility and rationality cognition for empathy still lacks comprehensive investigation. 

\subsection{Data Quality and Data Selection}
The advent of large language models has brought about a substantial increase in the volume of training data.~\cite{llama, openai2023gpt} In this scenario, the quality and quantity of data become paramount. LLMs, trained on vast amounts of data, can capture subtle nuances and complex patterns in language, excelling in various natural language processing tasks. However, the increase in data volume also brings new challenges, particularly in data management, cleaning, and annotation.~\cite{bai2024survey} In this section, we mainly discuss the effectiveness of data quality and data selection.

\paragraph{Data Quality}: High-quality data can significantly enhance the performance of models~\cite{llama3repo}. As the volume of data increases, ensuring high data quality becomes more challenging because it requires more resources for data cleaning, selection and annotation.~\cite{bai2024survey} Poor quality data can lead to models learning incorrect patterns and making inaccurate predictions. 

\paragraph{Data Selection}: 
LLMs-based methods were commonly used in data selection.~\cite{bai2024survey} For instance, \citet{du2023mods} leverages DeBERTa~\cite{he2020deberta} for scoring, retaining high-quality data, and combining it with the k-center greedy algorithm to select diverse data. \citet{chen2023alpagasus} score the accuracy of data using ChatGPT to pick out high-quality data. \citet{xu2023rethinking} use GPT-4 to rewrite data to increase their complexity and then streamline it by reducing its variety and improving its quality. \citet{liu2023makes} train two models using ChatGPT's labeled data to score the quality and complexity of the data. \citet{lu2023instag} rely on ChatGPT to tag each instance, defining its complexity and diversity based on these tags. \citet{parkar2024selectllm} first cluster the data, and then use GPT-4 to select high-quality data for each cluster.

Given the critical role of data quality and selection in enhancing model performance, our paper focuses on leveraging advanced data diversity and quality selection techniques to optimize empathetic data quality. By employing methods that integrate data features and similarity scores, we aim to efficiently identify and utilize high-quality data for empathetic response. 

\subsection{Data Generation}
Data has always been the key driver behind the success of large language models (LLMs). Recent advancements of LLMs are largely due to the availability of large-scale, diverse, and high-quality datasets for training these models ~\cite{privatesyn1} . However, the scarcity of data and the high costs present substantial challenges in obtaining such datasets ~\cite{privatesyn2, baize,cross-m}. Recent advancements in generating synthetic data and improving the performance of LLMs have shown promising results across various domains. Synthetic data holds great potential in building large-scale, high-quality datasets. Researchers have explored multiple approaches, from leveraging differential privacy to creating instruction-tuning frameworks, to enhance the quality, diversity, and utility of synthetic data ~\cite{generatedDataCL, Magicoder, MUFFIN, RefGPT}. A key component in generating high-quality synthetic datasets is precise alignment. Fan et al. ~\cite{reformattedalign} introduce REALIGN, a method that enhances the quality of instruction data by reformatting responses to better align with pre-established criteria and evidence, thereby improving LLMs' alignment with human values while minimizing human annotation and model hallucinations. Li et al. ~\cite{selfalignment} build a high-quality instruction-following language model by automatically labeling human-written text with corresponding instructions and demonstrating highly effective self-alignment. 

In the field of empathetic response, the construction of generative datasets remains relatively underexplored. Empathetic models primarily rely on the utilization and curation of existing datasets. Approaches such as those by ~\citet{sun2024efficient} have attempted to improve empathetic response performance through data selection methods. However, these methods still face challenges due to the lack of high-quality empathetic response data.
\section{Method}
In this section, we first introduce the empathetic data generation method in subsection \ref{sec:empathetic_generation}. Then we introduce the newly designed empathetic quality selection module in subsection \ref{sec:empathetic_quality_selection}. Additionally, we introduce the diversity selection in subsection \ref{sec:empathetic_diversity_selection}. We demonstrated the high-quality of our dataset in subsection \ref{sec:empathetic_high_quality}. With the curated dataset, we are able to fine-tune empathetic models.
\begin{figure*}
\centering 
\includegraphics[width=1.0\textwidth]{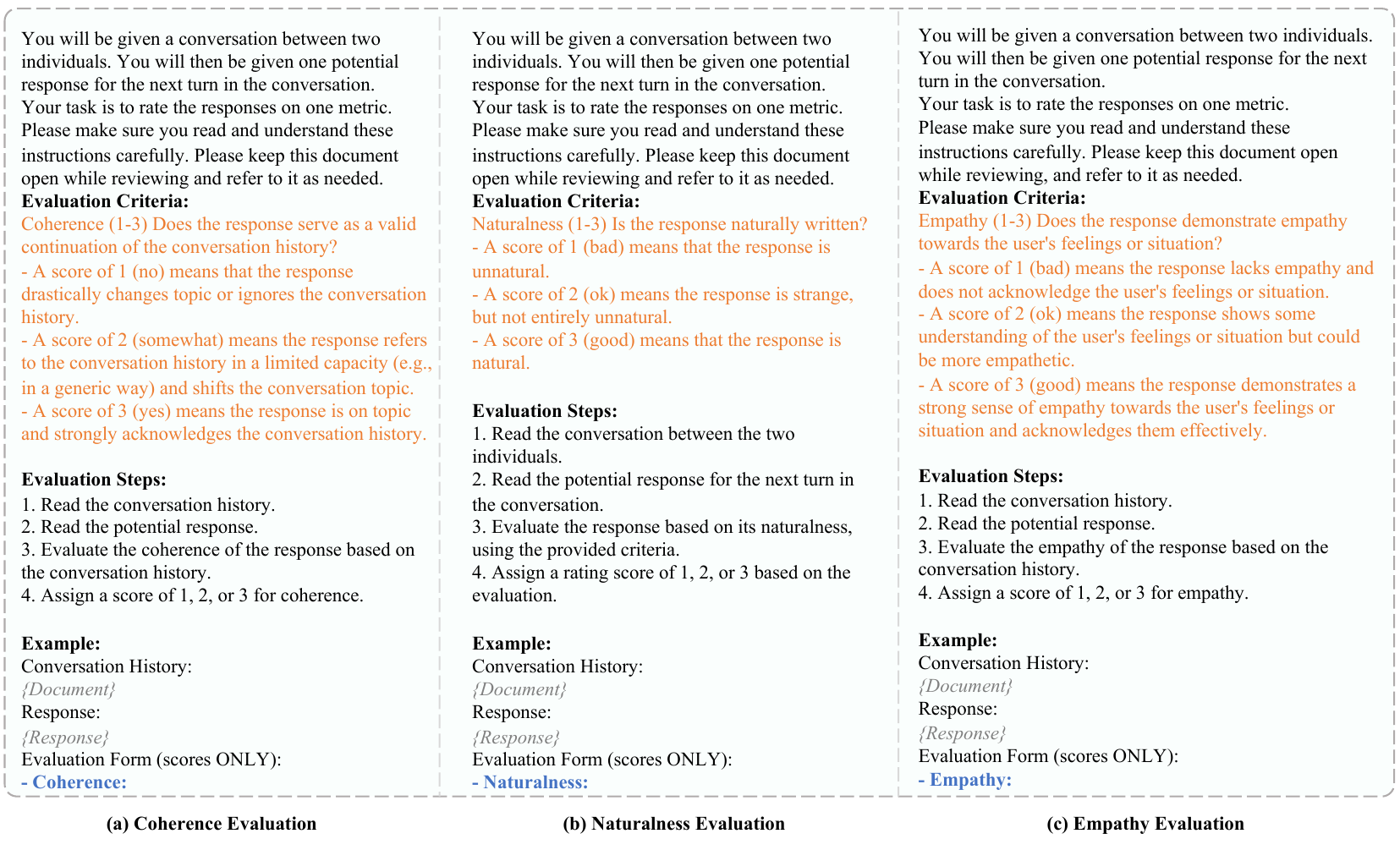} 
\caption{Data Quality Evaluation Prompts. (a) Assessing the coherence of the data. (b) Assessing the naturalness of the data. (c) Assessing the empathy of the data.}
\label{Fig.prompt}
\end{figure*}
\subsection{Empathetic Data Generation}\label{sec:empathetic_generation}

\begin{figure*}
\centering 
\includegraphics[width=1.0\textwidth]{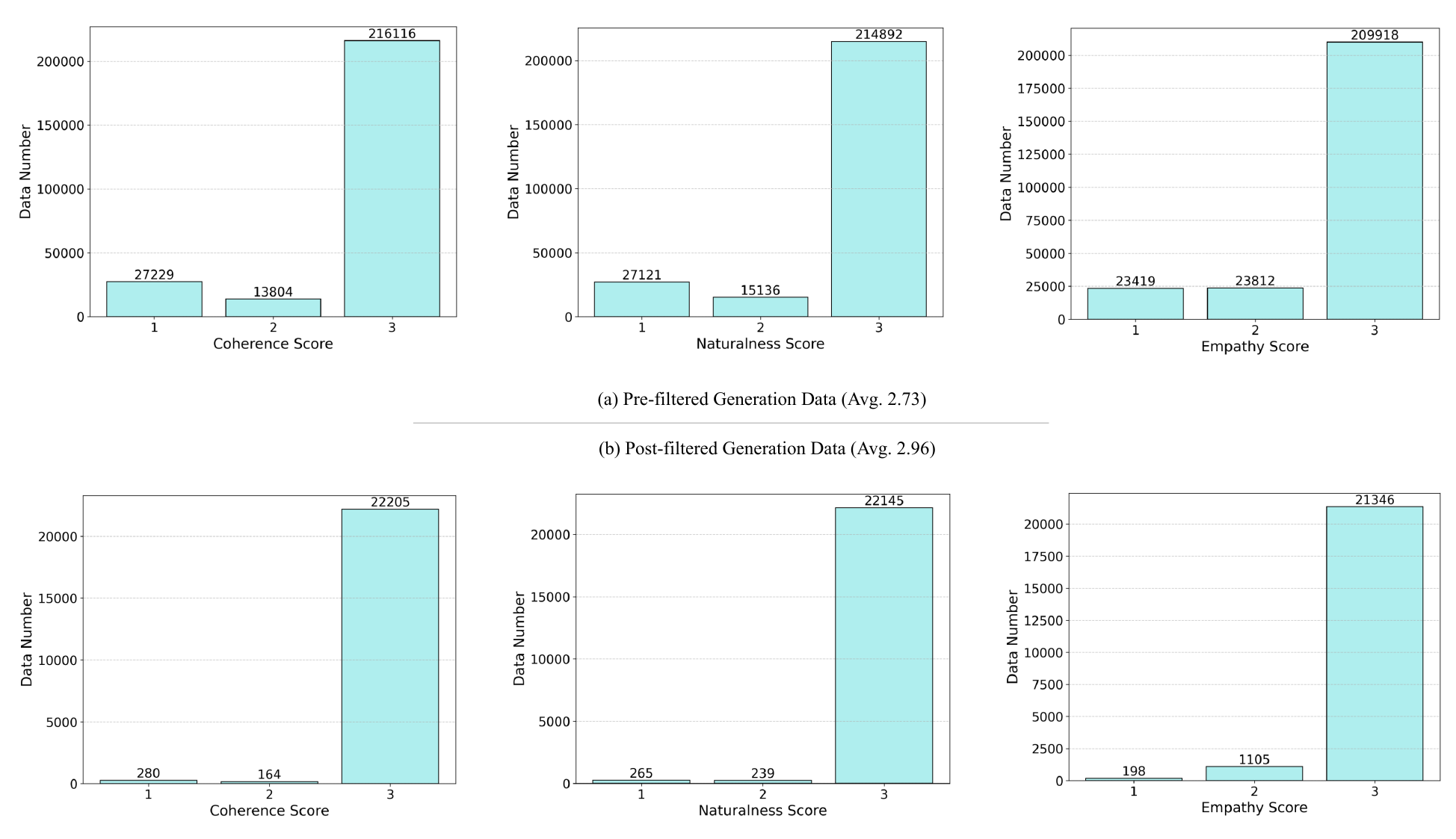} 
\caption{Scores of Coherence, Naturalness, and Empathy for Generated Data. (a) Scores before applying the data filtering strategy. (b) Scores after applying the data filtering strategy.}
\label{Fig.data-statistical}
\end{figure*}

\begin{figure}
\centering 
\includegraphics[width=0.49\textwidth]{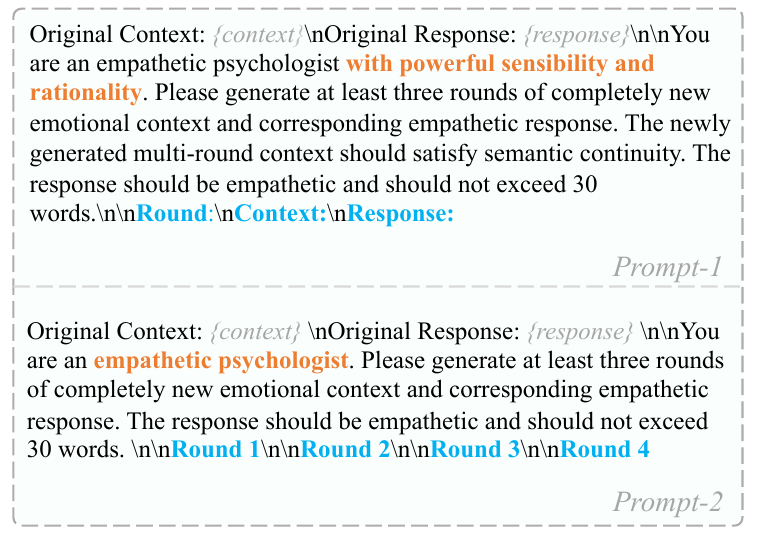} 
\caption{Prompts for generating empathetic responses.}
\label{Fig.prompt_generation}
\end{figure}

In this subsection, we introduce the data generation method. Our data generation is based on the authoritative EmpatheticDialogues (ED) dataset~\cite{rashkin2018towards}, which consists of 25,000 daily conversations encompassing 32 uniformly distributed emotional labels. By utilizing the prompt in Figure \ref{Fig.method}(1), we generate empathetic responses based on the high-quality ED dataset.

The formula for generating data using the prompt is as follows:

\begin{equation}
    \text{Synthetic Dataset} = \text{LLM}(\text{Prompt}(\text{ED Data}))
\end{equation}

Following the formula above, we generate more than 250k empathetic data points for further selection.
\subsection{Empathetic Data Quality Selection}\label{sec:empathetic_quality_selection}
In this subsection, we introduce the data quality selection process for empathetic data. Similar to subsection \ref{sec:empathetic_generation}, we utilize the ED dataset as the baseline and select data with an empathetic style similar to the ED dataset. The selection process is shown in Figure \ref{Fig.method}(2) and involves the following steps:

\paragraph{\textbf{Empathetic Discriminator}}
We fine-tune a LLM on the ED dataset to serve as the discriminator, denoted as $M_D$. This discriminator is trained to produce responses that mimic the high-quality empathetic style of the ED dataset, denoted as $D_{ED}$.
\begin{equation}
    M_D = \text{SFT}(\text{LLM}; D_{ED})
\end{equation}

\paragraph{\textbf{Similarity Based Quality Selection}}
In generating empathetic response data, we first use the discriminator $M_D$ to answer the question. Then, we use the SoTA embedding model gte-qwen2-7b-instruct to embed the model's response and the generated response. We calculate the similarity between the discriminator's response and the generated response. We only select the responses with similarity higher than the threshold $T$. The optimized procedure is as follows:
\begin{enumerate}
    \item Use the discriminator $M_D$ to answer the question, obtaining the response $r_D$:
    \[
    r_D = M_D(\text{question})
    \]

    \item Use gte-qwen2-7b-instruct to embed the discriminator's response $r_D$ and the generated response $r_G$, obtaining the embedding vectors $E_D$ and $E_G$:
    \[
    E_D = \text{gte-qwen2-7b-instruct}(r_D)
    \]
    \[
    E_G = \text{gte-qwen2-7b-instruct}(r_G)
    \]

    \item Calculate the similarity $S$ between the discriminator's response $r_D$ and the generated response $r_G$:
    \[
    S = \frac{E_D \cdot E_G}{\|E_D\| \|E_G\|}
    \]

    \item Select the generated response $r_G$ if the similarity $S$ is higher than the Domain Knowledge (DK) threshold $T$:
    \[
    r_G \text{ is selected if } S > T
    \]
\end{enumerate}

With the similarity selection based on the empathetic discriminator, we obtain a high-quality synthetic dataset.

\subsection{Empathetic Data Diversity Selection}\label{sec:empathetic_diversity_selection}
In this subsection, we introduce the diversity selection method. The aim of diversity selection is to maximize the minimum distance between selected data points, ensuring a diverse subset. The algorithm can be formally described as follows:

Given a dataset \(X = \{x_1, x_2, \ldots, x_n\}\), our goal is to select a subset \(S \subseteq X\) of size \(k\) that maximizes the minimum distance between any point in \(X\) and the closest point in \(S\). The objective function is:

\[
\max_{S \subseteq X, |S| = k} \min_{x \in X} \min_{s \in S} d(x, s)
\]

where \(d(x, s)\) denotes the distance between points \(x\) and \(s\).

As shown in Figure \ref{Fig.method}(3), we select the KCenterGreedy~\cite{sener2017active} algorithm for its robustness and quick computation. Using the KCenterGreedy algorithm, we obtain a high-diversity synthetic dataset.

\subsection{High Quality Generated Empathetic Data}\label{sec:empathetic_high_quality}
In this section, we introduce an analysis of the quality of our generated dataset. We use a three-dimensional evaluation to assess the quality of our curated data. Using the prompts in Figure \ref{Fig.prompt}, we evaluate the dataset's coherence, naturalness, and empathy, scoring each from 1 to 3 points. 

As shown in Figure \ref{Fig.data-statistical}, our method significantly improves the dataset's quality. The average score increased from 2.65 to 2.88. Additionally, more than 95\% of the data achieved full marks, demonstrating the high quality of our curated synthetic dataset. 

With high-quality generated empathetic response pairs, we can train a more effective empathetic model. Additionally, the empathetic response generation pipeline alleviates concerns about the quantity of empathetic data.


\section{Experiments}
In this section, we first introduce the experimental setups, including data selection and the training process. We then aim to answer the following questions to verify the effectiveness and robustness of our proposed Synth-Empathy model: \\
\textbf{Q1}: Can we fine-tune an empathetic model to achieve state-of-the-art (SoTA) performance with the curated synthetic data? \\
\textbf{Q2}: Can our model achieve favorable human evaluation results, given the importance of human evaluation for empathy? \\
\textbf{Q3}: Can we determine the trade-off between data quantity and quality to guide future empathetic data selection? \\
\textbf{Q4}: Are the quality selection and diversity selection modules necessary to enhance model performance?



\input{table/main}

\subsection{Experimental Setup}
\subsubsection{Datasets}

We evaluate our methods on the widely used EMPATHETICDIALOGUES dataset, which includes 25k multi-turn empathetic conversations.   Each conversation involves a speaker and a listener, with an average of 4.31 turns per dialog.   This dataset contains 32 evenly distributed emotion labels, with each conversation being associated with a specific label. And in order to guarantee the equity of the comparison baselines, we follow the same dataset division as the previous research method by splitting the dataset into training, validation, and test sets in an 8:1:1 ratio.

\subsubsection{Baselines}
The baseline models are listed as follows:

1. \textbf{SEEK}: Employ multi-granularity sentiment labels at the word, sentence, and dialogue levels to enhance emotional understanding.
2. \textbf{CASE}: Enhancing the cognitive capabilities of models with external knowledge sources, COMET and ConceptNet, to improve the quality of empathetic responses.
3. \textbf{ESCM}: Leveraging dynamic emotion-semantic vectors along with dependency trees to direct the model in generating empathetic responses.
4. \textbf{Qwen1.5-7B-F.T.}: Conducting SFT of Qwen1.5-7B-Chat model using ED dataset.
5. \textbf{LLaMA3-8B-F.T.}: Conducting SFT of LLaMA3-8B-Instruct model using ED dataset.
6. \textbf{LLaMA3-8B}: Generating empathetic response through designed prompt based on LLaMA3-8B-Instruct\footnote{\url{https://huggingface.co/meta-llama/Meta-Llama-3-8B-Instruct}} model.
7. \textbf{Mixtral-8x7B}: Generating empathetic response through designed prompt based on Mixtral-8x7B-Instruct\footnote{\url{https://huggingface.co/mistralai/Mixtral-8x7B-Instruct-v0.1}} model.
8. \textbf{Qwen1.5-72B-Chat}: Generating empathetic response through designed prompt based on Qwen1.5-72B-Chat\footnote{\url{https://huggingface.co/Qwen/Qwen1.5-72B-Chat}} model.

\subsubsection{Evaluation}

To offer a thorough evaluation of our model's capabilities, we assess its performance through both automatic metrics and human evaluations.

\textbf{Automatic Evaluation Metrics}: For automatic evaluation, we employ corpus-level BLEU (B-1 to B-4), sentence-level ROUGE (R-1, R-2), and Distinct (Dist-1, Dist-2) metrics. BLEU and ROUGE scores measure the similarity between the generated text and the reference text, with higher scores reflecting greater similarity. Meanwhile, Distinct-N assesses content diversity, where higher values indicate a broader range of diverse representations. CIDEr aligns more closely with human judgment of sentence similarity by employing TF-IDF to assign varying weights to different n-grams. Specifically, it reduces the weight of frequently occurring phrases while increasing the weight of less common ones, thereby reflecting the distinctiveness and significance of the n-grams in the evaluation process. Besides, we employed xxx to assess semantic similarity from the perspective of vector embedding.

\textbf{Human Evaluation Metrics}: For human evaluation, we choose the following metrics: Coherence, Empathy, Informativeness, and Continuity. \textbf{Coherence:} Assesses how well the model's generated text aligns with the intended response. \textbf{Empathy:} Measures the model's capacity to grasp the speaker's situation and convey appropriate concern. \textbf{Informativeness:} Evaluates the extent of information provided in the responses produced by the model. \textbf{Continuity:} Reflects the model's effectiveness in maintaining the flow of the conversation.

Then, the A/B test is conducted to evaluate the effectiveness of our model in comparison to several baseline models. Specifically, 200 examples are randomly selected from the test dataset. Each example is paired with two responses: one generated by our model and the other by a baseline model. Three evaluators assess each pair of responses, determining a winner, loser, or tie based on above four criteria.

\subsubsection{Settings}
All experiments are carried out on the machine equipped with 8 NVIDIA A100 GPUs, a 120-core CPU, and 960GB of memory. The inference process of LLMs is implemented on the Vllm framework with version 0.4.1.

\subsection{Main Experiments}
To address \textbf{Q1}, in this section, we evaluated various models on their performance in generating empathetic responses using several metrics, as summarized in Table~\ref{tab:main}. 

Our model was fine-tuned on the dataset combining the ED training set and our generated dataset. Our generated dataset comprises data with $S$ greater than 60 and diversity selection.

Our model consistently outperformed other models across most metrics. Specifically, it achieved the highest BLEU scores, with 22.05 for BLEU-1, 10.53 for BLEU-2, 5.99 for BLEU-3, and 3.77 for BLEU-4, demonstrating its superior ability to generate text closely aligned with reference translations. In terms of Distinct metrics, which gauge the diversity of generated responses, our model scored 3.23 for Distinct-1 and 18.84 for Distinct-2, indicating a strong balance between diversity and coherence.

Our model also excelled in the Rouge metrics, attaining scores of 18.72 for Rouge-1, 5.27 for Rouge-2, and 16.88 for Rouge-L. These scores reflect its effectiveness in capturing the essential content of the reference texts and producing more contextually relevant responses. In the CIDEr metric, which assesses content relevance and richness, our model achieved a score of 26.96, the highest among the compared models, highlighting its effectiveness in generating captioning that aligns closely with human judgments of content relevance and informativeness.

Furthermore, in embedding-based evaluations using BERT, our model demonstrated superior performance with a BERT-P score of 20.5, a BERT-R score of 18.2, and a BERT-F1 score of 19.4. These results indicate that our model generates text with more meaningful and contextually accurate semantic representations compared to other models. 

Overall, these results confirm that our model provides superior performance across a range of metrics, demonstrating its effectiveness in generating high-quality and diverse outputs. The promising results highlight the effectiveness of the high-quality synthetic dataset, thereby introducing a new paradigm for using synthetic data in empathetic responses.

\subsection{Human Evaluation}

\begin{figure}
\centering 
\includegraphics[width=0.5\textwidth]{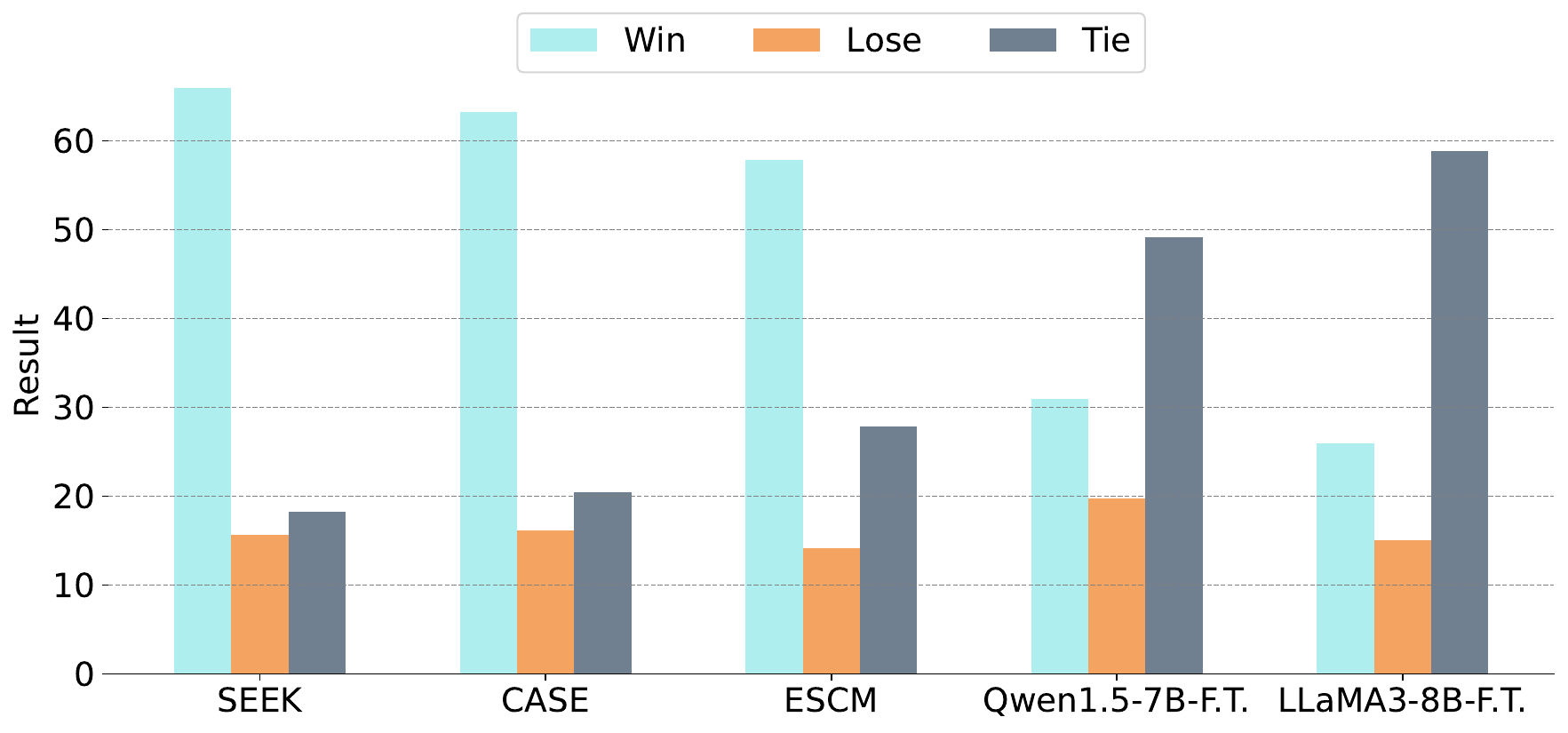} 
\caption{Human A/B test evaluation.}
\label{Fig.abtest}
\vspace{-4mm}
\end{figure}

To address \textbf{Q2}, we conducted a series of A/B tests to test the online performance. The comparative analysis includes our online model versus SEEK, CASE, ESCM, Qwen1.5-7B-F.T., and LLaMA3-8B-F.T. The results of these tests are summarized in Figure~\ref{Fig.abtest}.

The results of our A/B tests demonstrate that our model significantly outperforms SEEK, CASE, and ESCM, with win rates of 66\%, 63.3\%, and 57.9\%, respectively. When compared with the more advanced models, Qwen1.5-7B-F.T. and LLaMA3-8B-F.T., our model still demonstrates competitive performance. Against Qwen1.5-7B-F.T., our model secures a win rate of 31\% and maintains a reasonable loss rate of 19.8\%, with ties at 49.2\%. Similarly, our model achieves a win rate of 26\% against LLaMA3-8B-F.T., with a loss rate of 15.1\% and the highest tie rate of 58.9\%. 

Overall, our model's strong performance in human evaluations underscores its practicality and user-friendliness. These results indicate that our model is not only effective but also well-received by human evaluators, highlighting its potential for real-world applications.

\input{table/scale}
\subsection{Data Quality and Data Quantity}
\begin{figure}
\centering 
\includegraphics[width=0.47\textwidth]{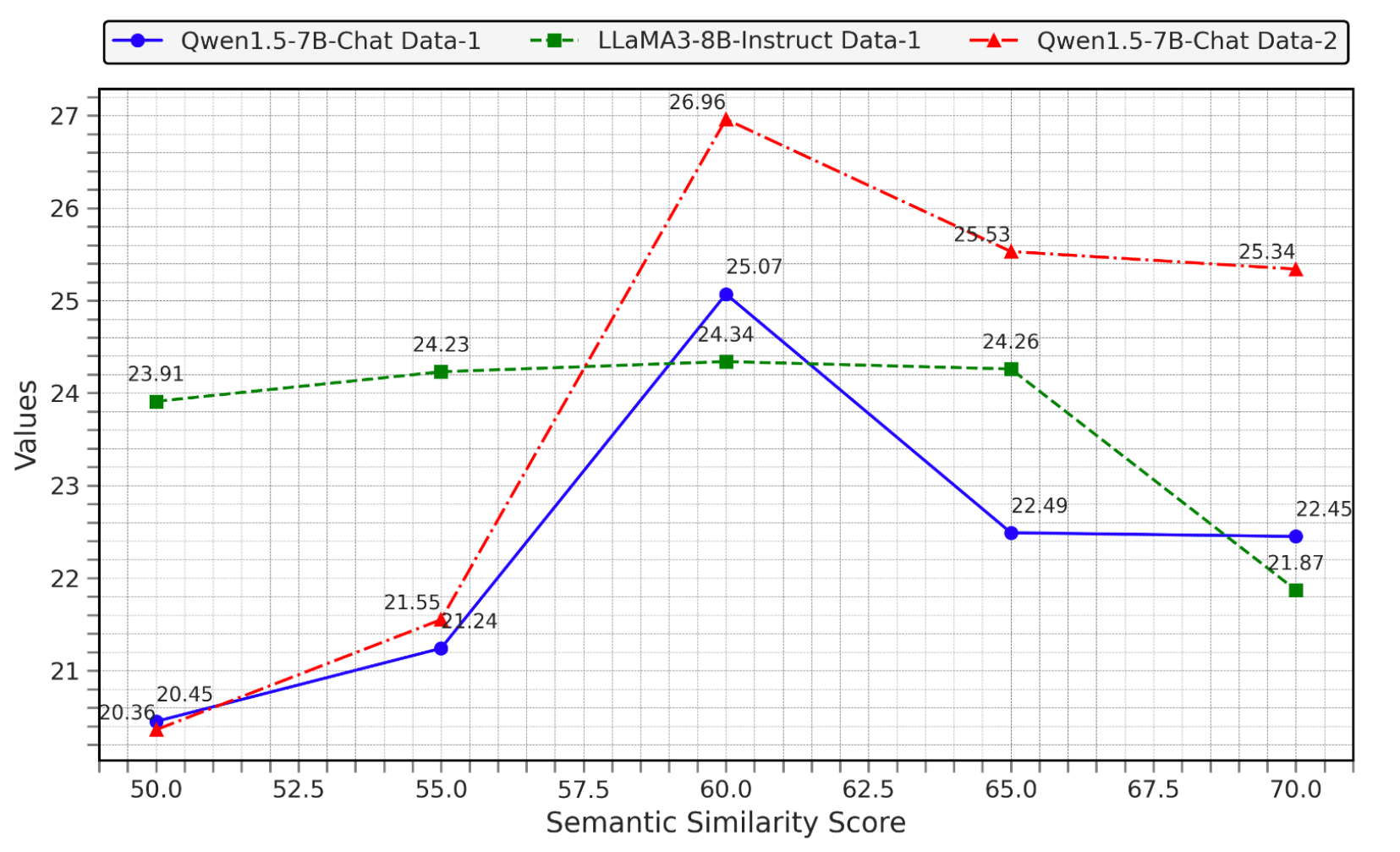} 
\caption{Visualization of trade-off between the DK ratio and the CIDEr score of generated empathetic responses.}
\label{Fig.scale-law}
\end{figure}

To address \textbf{Q3}, we analyzed the trade-off between the DK threshold, the similarity threshold in section \ref{sec:empathetic_quality_selection}, and the CIDEr score to demonstrate the robustness of our approach. Table~\ref{tab:scale} and Figure~\ref{Fig.scale-law} present detailed experimental results, comparing various DK thresholds across the Qwen1.5-7B and LLaMA3-8B models using two different data generation prompts.

As the DK threshold increases from 50 to 70, the CIDEr score of Qwen1.5-7B with Prompt 1 initially rises from 20.45 to a peak of 25.07 at a DK threshold of 60, before slightly decreasing to 22.45 at a DK threshold of 70. For the LLaMA3-8B model, the CIDEr score shows a similar pattern, peaking at 24.34 with a DK threshold of 60, before decreasing at higher DK ratios. In terms of Prompt 2, the Qwen1.5-7B model demonstrates an increase in the CIDEr score from 20.36 at a DK threshold of 50 to a peak of 26.96 at a DK threshold of 60, followed by a slight decline. This observation corroborates a clear trade-off trend, emphasizing that an optimal DK threshold enhances the generation of empathetic responses.

The consistency of the trade-off pattern across different models and prompts underscores the robustness of our method. Regardless of the specific model or prompt used, the CIDEr score consistently peaks around the DK threshold of 60, demonstrating that our method effectively balances data retention and response quality. Additionally, by analyzing the trade-off between data quantity and quality, we provide guidance for optimal empathetic selection methods in the future. Utilizing the best threshold can achieve superior empathetic performance and select the appropriate dataset, thereby avoiding the waste of computational resources.

\subsection{Ablation Study}
To address \textbf{Q4}, we provide the following ablation study. We remove the data quality selection module and the data diversity selection module to
demonstrate the effectiveness of our Synth-Empathy pipeline. 

\paragraph{Excluding Data Quality Selection Module.}
When excluding the data quality selection module, the performance of the model decreases across all benchmarks, as shown in Table \ref{tab:selection}. For both the Qwen1.5-7B and LLaMA-8B-Instruct models, this exclusion results in decreased performance, demonstrating the necessity of the quality selection module.

\paragraph{Excluding Data Diversity Selection Module.}
Similarly, the model's performance declines when the data diversity selection module is excluded, as shown in Table \ref{tab:diversity}. For both data selection ratios of 55 and 60, and with both the Qwen1.5-7B and LLaMA-8B-Instruct models, this exclusion results in decreased performance, demonstrating the necessity of the diversity selection module.

Overall, the ablation study highlights the critical role of data generation and data selection modules in SynthVLM. These experiments provide valuable insights into the contributions of each module, guiding future improvements and optimizations of the SynthVLM model.
\input{table/selection}

\input{table/diversity}

\section{Conclusion}
Empathy is a critical component of human social interaction and communication. However, high-quality empathetic data remains scarce, necessitating efficient and effective data generation algorithms. This paper presents Synth-Empathy, a new paradigm for generating high-quality empathetic data, pioneering a method to obtain empathetic responses without human labor. By leveraging quality and diversity selection, we achieved high-quality synthetic empathetic data. Remarkably, utilizing the curated empathetic data, our model outperforms all previous models. Additionally, our model achieved SoTA performance in human evaluation, demonstrating its contextual appropriateness and user-friendliness.
\newpage

\bibliographystyle{ACM-Reference-Format}
\bibliography{main}

\end{document}

%% file: table/main.tex
\begin{table*}[htbp]
  \centering
  \caption{Results of the automatic evaluation on ED test dataset and the
best performance are highlighted in bold. Our model trained on synthetic empathetic response data outperforms all other SoTA methods.}
  \resizebox{1.0\linewidth}{!}{
    \begin{tabular}{lccccccccccccc}
    \toprule
          & \multicolumn{4}{c}{\textbf{BLEU}} & \multicolumn{2}{c}{\textbf{Distinct}} & \multicolumn{3}{c}{\textbf{Rouge}} & \textbf{CIDEr} & \multicolumn{3}{c}{\textbf{Embedding}} \\
    \midrule
          & \textbf{B-1} & \textbf{B-2} & \textbf{B-3} & \textbf{B-4} & \textbf{D-1} & \textbf{D-2} & \textbf{R-1} & \textbf{R-2} & \textbf{R-l} & \textbf{CIDEr} & \textbf{BERT-P} & \textbf{BERT-R} & \textbf{BERT-F1} \\
    \midrule
    SEEK  & 10.77 & 4.40  & 2.02  & 1.08  & 0.68  & 2.81  & 12.74 & 2.94  & 12.07 & 9.95  & 19.9  & 8.2   & 14.0 \\
    CASE  & 15.21 & 6.97  & 3.75  & 2.27  & 0.78  & 4.34  & 16.14 & 4.38  & 14.99 & 19.11 & 20.5  & 14.1  & 17.3 \\
    ESCM  & 19.42 & 9.30  & 4.21  & 2.44  & 1.13  & 3.56  & 17.75 & 3.73  & 15.95 & 16.38 & 16.8  & 14.5  & 15.7 \\
    Qwen1.5-7B-F.T. & 22.00 & 10.31 & 5.70  & 3.53  & 3.19  & 19.48 & 18.12 & 4.87  & 16.35 & 22.04 & 20.1  & 17.8  & 19.0 \\
    LLaMA3-8B-F.T. & 21.28 & 9.69  & 5.18  & 3.11  & 4.02  & 24.00 & 17.93 & 4.76  & 16.07 & 24.02 & 20.1  & 17.9  & 19.0 \\
    LLaMA3-8B & 13.17 & 4.42  & 1.92  & 1.02  & 2.69  & 18.70 & 14.12 & 1.68  & 11.69 & 2.28  & 5.8   & 11.9  & 8.9 \\
    Mixtral-8x7B & 14.66 & 4.72  & 2.11  & 1.10  & 3.30  & 21.13 & 14.54 & 1.69  & 12.20 & 4.43  & 11.6  & 11.4  & 11.6 \\
    Qwen1.5-72B & 14.19 & 4.85  & 2.27  & 1.23  & 3.29  & 22.68 & 13.83 & 1.97  & 11.73 & 5.04  & 11.9  & 12.8  & 12.4 \\
    \textbf{ours} & \textbf{22.05} & \textbf{10.53} & \textbf{5.99} & \textbf{3.77} & 3.23  & 18.84 & \textbf{18.72} & \textbf{5.27} & \textbf{16.88} & \textbf{26.96} & \textbf{20.5} & \textbf{18.2} & \textbf{19.4} \\
    \bottomrule
    \end{tabular}%
    }
  \label{tab:main}%
\end{table*}%

%% file: table/scale.tex
\begin{table*}[htbp]
  \centering
  \caption{Trade-off between the DK ratio and
the quality of generated empathetic responses.}
  \resizebox{\linewidth}{!}{
    \begin{tabular}{cccccccccccc}
    \toprule
    \multirow{2}{*}{\textbf{Threshold}} & \multirow{2}{*}{\textbf{Models}} & \multirow{2}{*}{\textbf{Prompt}} & \multicolumn{4}{c}{\textbf{BLEU}} & \multicolumn{3}{c}{\textbf{Rouge}} & \multirow{2}{*}{\textbf{CIDEr}} & \multirow{2}{*}{\textbf{Data}} \\
    \cmidrule(lr){4-7} \cmidrule(lr){8-10}
          &   &   & \textbf{B-1} & \textbf{B-2} & \textbf{B-3} & \textbf{B-4} & \textbf{R-1} & \textbf{R-2} & \textbf{R-l} &   &   \\
    \midrule
    50 & \multirow{5}{*}{Qwen1.5-7B} & \multirow{5}{*}{Prompt1} & 21.31 & 9.52  & 5.12  & 3.10  & 17.66 & 4.61  & 15.95 & 20.45 & 115,898 \\
    55 &  &  & 20.97 & 9.39  & 5.20  & 3.21  & 17.75 & 4.59  & 15.93 & 21.24 & 77,182 \\
    60 &  &  & 22.48 & 10.47 & 5.76  & 3.54  & 19.04 & 5.21  & 17.16 & 25.07 & 45,298 \\
    65 &  &  & 22.04 & 10.26 & 5.72  & 3.53  & 18.13 & 4.95  & 16.40 & 22.49 & 22,705 \\
    70 &  &  & 21.60 & 9.77  & 5.32  & 3.27  & 18.20 & 4.880 & 16.32 & 22.45 & 9,791 \\
    \midrule
    50 & \multirow{5}{*}{LLaMA3-8B-Instruct} & \multirow{5}{*}{Prompt1} & 20.25 & 9.28  & 5.07  & 3.10  & 17.62 & 4.60  & 15.98 & 23.91 & 115,898 \\
    55 &  &  & 20.47 & 9.35  & 5.08  & 3.07  & 17.64 & 4.60  & 15.96 & 24.23 & 77,182 \\
    60 &  & & 21.35 & 9.87  & 5.42  & 3.33  & 17.98 & 4.84  & 16.25 & 24.34 & 45,298 \\
    65 &  & & 20.08 & 9.23  & 5.05  & 3.09  & 17.53 & 4.78  & 15.95 & 24.26 & 22,705 \\
    70 &  & & 20.53 & 9.15  & 4.93  & 2.99  & 17.20 & 4.22  & 15.45 & 21.87 & 9,791 \\
    \midrule
    50 & \multirow{5}{*}{Qwen1.5-7B} & \multirow{5}{*}{Prompt2} & 20.59 & 9.17  & 4.97  & 3.01  & 17.38 & 4.44  & 15.53 & 20.36 & 64,020 \\
    55 &  &  & 21.61 & 9.80  & 5.37  & 3.28  & 17.82 & 4.71  & 16.01 & 21.55 & 37,043 \\
    60 &  &  & 22.05 & 10.53 & 5.99  & 3.77  & 18.72 & 5.27  & 16.88 & 26.96 & 18,789 \\
    65 &  &  & 22.80 & 10.77 & 6.04  & 3.74  & 18.81 & 5.15  & 17.00 & 25.53 & 8,227 \\
    70 &  &  & 22.69 & 10.65 & 5.86  & 3.58  & 19.02 & 5.23  & 17.20 & 25.34 & 3,041 \\
    \bottomrule
    \end{tabular}%
    }
  \label{tab:scale}%
  \vspace{-2mm}
\end{table*}%

%% file: table/selection.tex
\begin{table}[htbp]
  \centering
  \caption{Ablation study on quality selection. Excluding the data quality selection module results in a performance drop.}
  \resizebox{\linewidth}{!}{
    \begin{tabular}{ccccccccc}
    \toprule
    \textbf{Threshold} & \textbf{Model} & \textbf{Quality} & \textbf{B-1} & \textbf{B-2} & \textbf{B-3} & \textbf{B-4} & \textbf{R-2} & \textbf{CIDEr} \\
    \midrule
    \multirow{2}{*}{60} & \multirow{2}{*}{Qwen1.5-7B} & \cmark & 22.48 & 10.47 & 5.76  & 3.54  & 5.21  & 25.07 \\
          &  & \xmark & 22.00\textcolor{blue}{$\downarrow$} & 10.31\textcolor{blue}{$\downarrow$} & 5.70\textcolor{blue}{$\downarrow$}  & 3.53\textcolor{blue}{$\downarrow$}  & 4.87\textcolor{blue}{$\downarrow$}  & 22.04\textcolor{blue}{$\downarrow$} \\
    \midrule
    \multirow{2}{*}{60} & \multirow{2}{*}{LLaMA3-8B-Instruct} & \cmark & 21.35 & 9.87  & 5.42  & 3.33  & 4.84  & 24.34 \\
          &  & \xmark & 21.28\textcolor{blue}{$\downarrow$} & 9.69\textcolor{blue}{$\downarrow$}  & 5.18\textcolor{blue}{$\downarrow$}  & 3.11\textcolor{blue}{$\downarrow$}  & 4.76\textcolor{blue}{$\downarrow$}  & 24.02\textcolor{blue}{$\downarrow$} \\
    \bottomrule
    \end{tabular}%
    }
  \label{tab:selection}%
\end{table}%


%% file: table/diversity.tex
\begin{table}[htbp]
  \centering
  \caption{Ablation study on diversity selection. Excluding data diversity selection module results in a performance drop.}
  \resizebox{\linewidth}{!}{
    \begin{tabular}{ccccccccc}
    \toprule
    \textbf{Threshold} & \textbf{Model} & \textbf{Diversity} & \textbf{B-1} & \textbf{B-2} & \textbf{B-3} & \textbf{B-4} & \textbf{R-2} & \textbf{CIDEr} \\
    \midrule
    \multirow{2}{*}{55} & \multirow{2}{*}{Qwen1.5-7B} & \cmark & 21.01 & 9.55  & 5.26  & 3.24  & 4.65  & 22.60 \\
          &  & \xmark & 20.97\textcolor{blue}{$\downarrow$} & 9.39\textcolor{blue}{$\downarrow$}  & 5.20\textcolor{blue}{$\downarrow$}  & 3.21\textcolor{blue}{$\downarrow$}  & 4.59\textcolor{blue}{$\downarrow$}  & 21.24\textcolor{blue}{$\downarrow$} \\
    \midrule
    \multirow{2}{*}{60} & \multirow{2}{*}{Qwen1.5-7B} & \cmark & 22.23 & 10.66 & 6.02  & 3.77  & 5.30  & 26.52 \\
          &  & \xmark & 22.48 & 10.47\textcolor{blue}{$\downarrow$} & 5.76\textcolor{blue}{$\downarrow$}  & 3.54\textcolor{blue}{$\downarrow$}  & 5.21\textcolor{blue}{$\downarrow$}  & 25.07\textcolor{blue}{$\downarrow$} \\
    \midrule
    \multirow{2}{*}{55} & \multirow{2}{*}{LLaMA3-8B-Instruct} & \cmark & 21.83 & 9.92  & 5.41  & 3.30  & 4.78  & 24.34 \\
          &  & \xmark & 20.47\textcolor{blue}{$\downarrow$} & 9.35\textcolor{blue}{$\downarrow$}  & 5.08\textcolor{blue}{$\downarrow$}  & 3.07\textcolor{blue}{$\downarrow$}  & 4.60\textcolor{blue}{$\downarrow$}  & 24.23\textcolor{blue}{$\downarrow$} \\
    \midrule
    \multirow{2}{*}{60} & \multirow{2}{*}{LLaMA3-8B-Instruct} & \cmark & 21.94 & 10.08 & 5.47  & 3.34  & 4.91  & 25.22 \\
          &  & \xmark & 21.35\textcolor{blue}{$\downarrow$} & 9.87\textcolor{blue}{$\downarrow$}  & 5.42\textcolor{blue}{$\downarrow$}  & 3.33\textcolor{blue}{$\downarrow$}  & 4.84\textcolor{blue}{$\downarrow$}  & 24.34\textcolor{blue}{$\downarrow$} \\
    \bottomrule
    \end{tabular}%
    }
  \label{tab:diversity}%
  \vspace{-2mm}
\end{table}%